\documentclass[runningheads]{llncs}
\usepackage{cite}
\usepackage{amsmath,amssymb,amsfonts}
\usepackage{algorithmic}
\usepackage{graphicx}
\usepackage{textcomp}
\usepackage{xcolor}
\usepackage{url}
\usepackage{multirow}
\usepackage{booktabs}
\usepackage{verbatim}
\usepackage{bbm}
\usepackage{marvosym}
\usepackage{ifsym}

\begin{document}
\title{Contrastive Learning-based Imputation-Prediction Networks for In-hospital Mortality Risk Modeling using EHRs}
\author{Yuxi Liu\Letter\inst{1} \and Zhenhao Zhang\inst{2} \and Shaowen Qin\inst{1} \and Flora D. Salim\inst{3} \and Antonio Jimeno Yepes\inst{4}}
\toctitle{Contrastive Learning-based Imputation-Prediction Networks for In-hospital Mortality Risk Modeling using EHRs}
\tocauthor{Yuxi Liu, Zhenhao Zhang, Shaowen Qin, Flora D. Salim, Antonio Jimeno Yepes}
\institute{College of Science and Engineering, Flinders University, Tonsley, SA 5042, Australia
\email{\{liu1356, shaowen.qin\}@flinders.edu.au} \\ \and
College of Life Sciences, Northwest A\&F University, Yangling, Shaanxi 712100, China
\email{\{zhangzhenhow\}@nwafu.edu.cn} \\ \and
School of Computer Science and Engineering, UNSW, Sydney, NSW 2052, Australia
\email{\{flora.salim\}@unsw.edu.au} \\ \and
School of Computing Technologies, RMIT University, Melbourne, VIC 3001, Australia
\email{\{antonio.jose.jimeno.yepes\}@rmit.edu.au} \\
}
\pagestyle{empty}
\thispagestyle{empty}
\maketitle              
\begin{abstract}
Predicting the risk of in-hospital mortality from electronic health records (EHRs) has received considerable attention. Such predictions will provide early warning of a patient's health condition to healthcare professionals so that timely interventions can be taken. This prediction task is challenging since EHR data are intrinsically irregular, with not only many missing values but also varying time intervals between medical records. Existing approaches focus on exploiting the variable correlations in patient medical records to impute missing values and establishing time-decay mechanisms to deal with such irregularity. This paper presents a novel contrastive learning-based imputation-prediction network for predicting in-hospital mortality risks using EHR data. Our approach introduces graph analysis-based patient stratification modeling in the imputation process to group similar patients. This allows information of similar patients only to be used, in addition to personal contextual information, for missing value imputation. Moreover, our approach can integrate contrastive learning into the proposed network architecture to enhance patient representation learning and predictive performance on the classification task. Experiments on two real-world EHR datasets show that our approach outperforms the state-of-the-art approaches in both imputation and prediction tasks.

\end{abstract}

\section{Introduction}
The broad adoption of digital healthcare systems produces a large amount of electronic health records (EHRs) data, providing us the possibility to develop predictive models and tools using machine learning techniques that would enable healthcare professionals to make better decisions and improve healthcare outcomes. One of the EHR-based risk prediction tasks is to predict the mortality risk of patients based on their historical EHR data \cite{harutyunyan2019multitask,sheikhalishahi2020benchmarking}. The predicted mortality risks can be used to provide early warnings when a patient's health condition is about to deteriorate so that more proactive interventions can be taken.

\begin{figure*}[!htb]
        \centering
        \includegraphics[width = 1.0\linewidth]{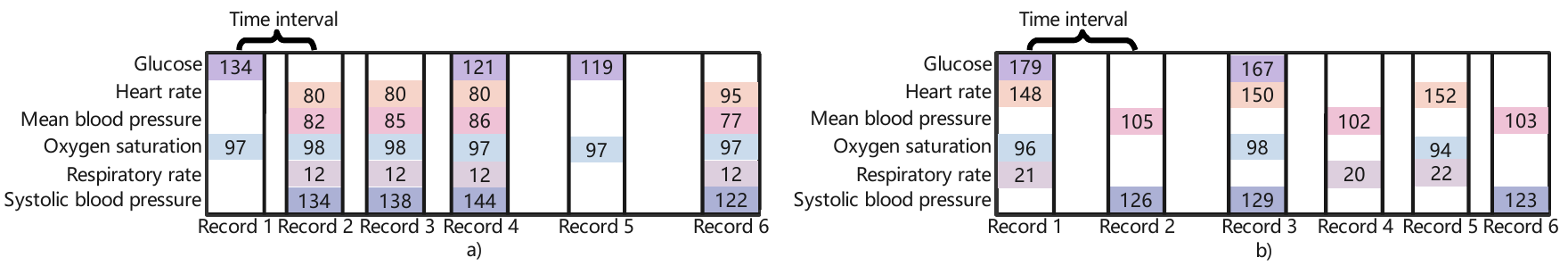}
        \caption{\textbf{Illustration of medical records of patients A and B.}}
        \label{fig:DATAFRAME}
\end{figure*}

However, due to a high degree of irregularity in the raw EHR data, it is challenging to directly apply traditional machine learning techniques to perform predictive modeling. We take the medical records of two anonymous patients from the publicly available MIMIC-III database and present these in Figure~\ref{fig:DATAFRAME} as an example. Figure~\ref{fig:DATAFRAME} clearly indicates the irregularity problem, including many missing values and varying time intervals between medical records.

Most studies have focused on exploiting variable correlations in patient medical records to impute missing values and establishing time-decay mechanisms to take into account the effect of varying time intervals between records \cite{oh2021sting,che2018recurrent,cao2018brits,luo2018multivariate,luo2019e2gan,tan2020data,mulyadi2021uncertainty,ni2022mbgan}. After obtaining the complete data matrices from the imputation task, the complete data matrices are used as input for downstream healthcare prediction tasks \cite{che2018recurrent,cao2018brits,luo2018multivariate,luo2019e2gan,tan2020data,mulyadi2021uncertainty,xu2020deep,shi2021deep,mccombe2021practical,pereira2022partial,lee2022multi}. Although these studies have achieved satisfactory imputation performance, consideration of using the information of similar patients on the imputation task, which might lead to improved imputation performance, has not yet been fully experimented. Furthermore, with imputation data, high-quality representation must be applied, as the imputation data may affect the performance of downstream healthcare prediction tasks.

Patient stratification refers to the method of dividing a patient population into subgroups based on specific disease characteristics and symptom severity. Patients in the same subgroup generally had more similar health trajectories. Therefore, we propose to impute missing values in patient data using information from the subgroup of similar patients rather than the entire patient population.

In this paper, we propose a novel contrastive learning-based imputation-prediction network with the aim of improving in-hospital mortality prediction performance using EHR data. Missing value imputation for EHR data is done by exploiting similar patient information as well as patients' personal contextual information. Similar patients are generated from patient similarity calculation during stratification modeling and analysis of patient graphs.

Contrastive learning has been proven to be an important machine learning technique in the computer vision community \cite{le2020contrastive}. In contrastive learning, representations are learned by comparing input samples. The comparisons are made on the similarity between positive pairs or dissimilarity between negative pairs. The main goal is to learn an embedding space where similar samples are put closer to each other while dissimilar samples are pushed farther apart. Contrastive learning can be applied in both supervised \cite{khosla2020supervised,wang2021exploring,zang2021scehr} and unsupervised \cite{li2022cluster,li2022uctopic,pang2022unsupervised} settings.

Motivated by the recent developments in contrastive representation learning \cite{yang2022mutual,yuan2021multimodal,wang2021multi}, we integrate contrastive learning into the proposed network architecture to perform imputation and prediction tasks. The benefit of incorporating contrastive learning into the imputation task is that such an approach can enhance patient representation learning by keeping patients of the same stratification together and pushing away patients from different stratifications. This would lead to enhanced imputation performance. The benefit of incorporating contrastive learning into the prediction task is improved predictive performance of the binary classification problem (i.e., the risk of death and no death), which is achieved by keeping the instances of a positive class closer and pushing away instances from a negative class.

Our major contributions are as follows:
\begin{itemize}
    \item To the best of our knowledge, this is the first attempt to consider patient similarity via stratification of EHR data on the imputation task.
    \item We propose a novel imputation-prediction approach to perform imputation and prediction simultaneously with EHR data.
    \item We successfully integrate contrastive learning into the proposed network architecture to improve imputation and prediction performance.
    \item Extensive experiments conducted on two real-world EHR datasets show that our approach outperforms all baseline approaches in imputation and prediction tasks.
\end{itemize}

\section{Related Work}
There has been an increased interest in EHR-based health risk predictions \cite{cui2022automed,li2021integrating,ma2020concare,ma2020adacare,ma2021distilling}. It has been recognized that EHR data often contains many missing values due to patient conditions and treatment decisions \cite{tan2020data}. Existing research addresses this challenge by imputing missing data and feeding them into the supervised algorithms as auxiliary information \cite{groenwold2020informative}. GRU-D \cite{che2018recurrent} represents such an example. The GRU-D is built upon the Gated Recurrent Unit \cite{cho2014learning}. GRU-D proposes to impute missing values by decaying the contributions of previous observation values toward the overall mean over time. Similarly, BRITS \cite{cao2018brits} incorporates a bidirectional recurrent neural network (RNN) to impute missing values. Since the incorporated bidirectional RNN learns EHR data in both forward and backward directions, the accumulated loss is introduced to train the model.

Another line of related work is based on the generative adversarial network (GAN) architecture, which aims at treating the problem of missing data imputation as data generation. The intuitions behind GAN can be seen as making a generator and a discriminator against each other \cite{goodfellow2020generative}. The generator generates fake samples from random 'noise' vectors, and the discriminator distinguishes the generator's fake samples from actual samples. Examples of research into GAN-based imputation methods include GRUI-GAN \cite{luo2018multivariate}, E$^{2}$GAN \cite{luo2019e2gan}, E$^{2}$GAN-RF \cite{zhang2021missing}, and STING \cite{oh2021sting}. These studies take the vector of actual samples, which has many missing values, use a generator to generate the corresponding imputed values and distinguish the generated imputed values from real values using a discriminator.

Several studies have evaluated the effectiveness of applying transformer-based imputation methods to EHR data. Examples of representative studies include MTSIT \cite{yildiz2022multivariate} and MIAM \cite{lee2022multi}. The MTSIT is built with an autoencoder architecture to perform missing value imputation in an unsupervised manner. The autoencoder architecture used in MTSIT includes the Transformer encoder \cite{vaswani2017attention} and a linear decoder, which are implemented with a joint reconstruction and imputation approach. The MIAM is built upon the self-attention mechanism \cite{vaswani2017attention}. Given EHR data, MIAM imputes the missing values by extracting the relationship among the observed values, missingness indicators (0 for missing and 1 for not missing), and the time interval between consecutive observations.

\section{Method}
\subsection{Network Architecture}
\begin{figure*}[!htbp]
        \centering
        \includegraphics[width = 1.0\linewidth]{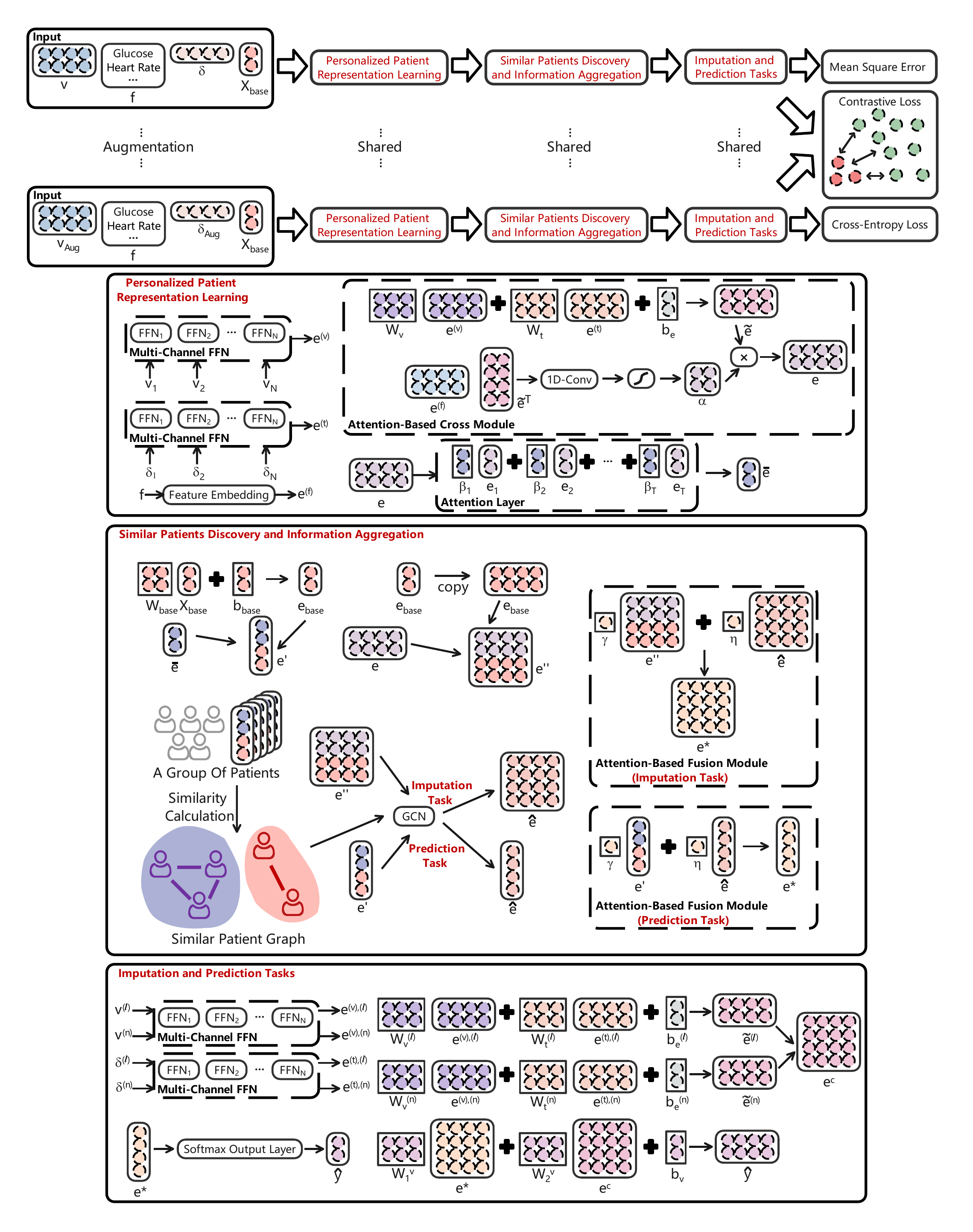}
        \caption{\textbf{Schematic description of the proposed network.}}
        \label{fig:OVERVIEW}
\end{figure*}

The architecture of the proposed network is shown in Figure~\ref{fig:OVERVIEW}.
\subsubsection{Data Representation}
We represent a multivariate time series $X$ with up to $N$ variables of length $T$ as a set of observed triplets, i.e., $X = \{(f_{i}, v_{i}, t_{i})\}^{N}_{i = 1}$. An observed triplet is represented as a $(f, v, t)$, where $f \in F$ is the variable/feature, $v \in \mathbb{R}^{T}$ is the observed value, and $t \in \mathbb{R}^{T}$ is the time. We incorporate a masking vector $m_{i}$ to represent missing values in $v_{i}$ as:
\begin{equation}
\begin{split}
\label{eq:1}
m_{i, t} =
  \begin{cases}
    1,  & if\ v_{i, t}\ is\ observed \\
    0,  & otherwise
  \end{cases}
\end{split}
\end{equation}

Let $\delta \in \mathbb{R}^{N \times T}$, $\delta^{(l)} \in \mathbb{R}^{N \times T}$, and $\delta^{(n)} \in \mathbb{R}^{N \times T}$ denote three time interval matrices. $\delta_{t}$ is the time interval between the current time $t$ and the last time $t-1$. $\delta^{(l)}_{i, t}$ is the time interval between the current time $t$ and the time where the i-th variable is observed the last time. $\delta^{(n)}_{i, t}$ is the time interval between the current time $t$ and the time where the i-th variable is observed next time. $\delta^{(l)}_{i, t}$ and $\delta^{(n)}_{i, t}$ can be written as:
\begin{equation}
\begin{split}
\label{eq:2}
\delta^{(l)}_{i, t} =
  \begin{cases}
   \delta_{i, t},  & if\ m_{i, t - 1} = 1 \\
   \delta_{i, t} + \delta^{(l)}_{i, t - 1},  & otherwise
  \end{cases}
\end{split}
\end{equation}

\begin{equation}
\begin{split}
\label{eq:3}
\delta^{(n)}_{i, t} =
  \begin{cases}
   \delta_{i, t + 1},  & if\ m_{i, t + 1} = 1 \\
   \delta_{i, t + 1} + \delta^{(n)}_{i, t + 1},  & otherwise
  \end{cases}
\end{split}
\end{equation}

Let $v^{(l)}$ and $v^{(n)}$ denote two neighboring value matrices, the observed values of the last time and next time. $v^{(l)}$ and $v^{(n)}$ can be written as: \begin{equation}
\begin{split}
\label{eq:4}
v^{(l)}_{i, t} =
  \begin{cases}
   v_{i, t - 1},  & if\ m_{i, t - 1} = 1 \\
   v^{(l)}_{i, t - 1},  & otherwise
  \end{cases}
\end{split}
\end{equation}

\begin{equation}
\begin{split}
\label{eq:5}
v^{(n)}_{i, t} =
  \begin{cases}
   v_{i, t + 1},  & if\ m_{i, t + 1} = 1 \\
   v^{(n)}_{i, t + 1},  & otherwise
  \end{cases}
\end{split}
\end{equation}
where $v^{(l)}_{i, t}$ and $v^{(n)}_{i, t}$ are the values of the i-th variable of $v_{t}^{(l)}$ and $v_{t}^{(n)}$.

Let $D = \{(X_{p}, y_{p})\}^{P}_{p = 1}$ denote the EHR dataset with up to $P$ labeled samples. The p-th sample contains a multivariate time series $X_{p}$ consisting of the physiological variables, and a binary label of in-hospital mortality $y_{p} \in \{0, 1\}$. Let $X_{base} \in \mathbb{R}^{g}$ denote the patient-specific characteristics (i.e., age, sex, ethnicity, admission diagnosis) with up to $g$ dimension.

\subsubsection{Personalized Patient Representation Learning}
Given an input multivariate time series/a single patient data $X = \{(f_{i}, v_{i}, t_{i})\}^{N}_{i = 1}$, the embedding for the i-th triplet $e_{i} \in \mathbb{R}^{d}$ is generated by aggregating the feature embedding $e_{i}^{(f)} \in \mathbb{R}^{d}$, the value embedding $e_{i}^{(v)} \in \mathbb{R}^{d \times T}$, and the time interval embedding $e^{(t)}_{i} \in \mathbb{R}^{d \times T}$. The feature embedding is similar to the word embedding, which allows features with similar meanings to have a similar representation. Particularly, the value embedding and time interval embedding are obtained by separately implementing a multi-channel feed-forward neural network (FFN) as:
\begin{equation}
\begin{split}
\label{eq:6}
e_{i, 1}^{(v)}, \cdots, e_{i, T}^{(v)} = FFN_{i}^{(v)} (v_{i, 1}, \cdots, v_{i, T}), \\
e_{i, 1}^{(t)}, \cdots, e_{i, T}^{(t)} = FFN_{i}^{(t)} (\delta_{i, 1}, \cdots, \delta_{i, T}).
\end{split}
\end{equation}

Through the processes above, we are able to obtain $e^{(f)} \in \mathbb{R}^{Nd}$, $e^{(v)} \in \mathbb{R}^{Nd \times T}$, and $e^{(t)} \in \mathbb{R}^{Nd \times T}$, which are fed into the attention-based cross module to generate an overall representation. Note that $e^{(f)} \in \mathbb{R}^{Nd}$ is expanded into $e^{(f)} \in \mathbb{R}^{Nd \times T}$. Specifically, we design the attention-based cross module to generate a cross-attention matrix as:
\begin{equation}
\begin{split}
\label{eq:7}
\tilde{e} = W_{v} \cdot e^{(v)} + W_{t} \cdot e^{(t)} + b_{e}, \\
E = ScaledDot(e^{(f)}, \tilde{e}) = \frac{e^{(f)} \cdot \tilde{e}^{\top}}{\sqrt{d}},
\end{split}
\end{equation}
where $E \in \mathbb{R}^{Nd \times Nd }$ is the cross-attention matrix that corresponds to the scaled-dot similarity. We then apply a 1D convolutional layer to the cross-attention matrix $E$ as:
\begin{equation}
\begin{split}
\label{eq:8}
\alpha = Softmax(Conv(E)),
\end{split}
\end{equation}
where $Conv$ is the 1D convolutional layer and $\alpha$ is the cross-attention score matrix. We integrate $\alpha$ and $\tilde{e}$ into a weighted representation $e$ as:
\begin{equation}
\begin{split}
\label{eq:9}
e = \alpha \odot \tilde{e}.
\end{split}
\end{equation}

Given a batch of patients, the embedding for them can be written as:
\begin{equation}
\begin{split}
\label{eq:10}
e = [e_{1}, e_{2}, \cdots, e_{B}] \in \mathbb{R}^{B \times Nd \times T},
\end{split}
\end{equation}
where $B$ is the batch size. Since $e$ still takes the form of sequence data, we design an attention layer to generate a series of attention weights $(\beta_{1}, \beta_{2}, \cdots, \beta_{T})$ and reweight these weights to produce an overall feature representation as:
\begin{equation}
\begin{split}
\label{eq:11}
\beta = Softmax(e \cdot W_{e} + b_{e}), \\
\bar{e} = \sum^{T}_{t = 1} \beta_{t} \odot e_{t},
\end{split}
\end{equation}
where $\bar{e} \in \mathbb{R}^{B \times Nd}$ is the new generated patient representation.

\subsubsection{Similar Patients Discovery and Information Aggregation}
Before conducting patient similarity calculation, we encode $X_{base} \in \mathbb{R}^{g}$ as $e_{base} \in \mathbb{R}^{d_{g}}$ and concatenate $e_{base}$ with $\bar{e}$ as:
\begin{equation}
\begin{split}
\label{eq:12}
e_{base} =  W_{base} \cdot X_{base} + b_{base}, \\
e^{\prime} = Concate(\bar{e}, e_{base}),
\end{split}
\end{equation}
where Concate is the concatenation operation.

For the batch of patient representations, the pairwise similarities that correspond to any two patient representations can be calculated as:
\begin{equation}
\begin{split}
\label{eq:13}
\Lambda = sim(e^{\prime}, e^{\prime}) = \frac{e^{\prime} \cdot e^{\prime}}{(Nd+d_{g})^{2}},
\end{split}
\end{equation}
where $sim(\cdot)$ is the measure of cosine similarity and $\Lambda \in \mathbb{R}^{B \times B}$ is the patient similarity matrix.

Moreover, we incorporate a learnable threshold $\varphi$ into the patient similarity calculation to filter out similarities below the threshold. The similarity matrix can be rewritten as:
\begin{equation}
\begin{split}
\label{eq:14}
\Lambda^{\prime} =
  \begin{cases}
    \Lambda,  & if\ \Lambda\ >\ \varphi \\
    0,  & otherwise
  \end{cases}
\end{split}
\end{equation}

We take into account the batch of patients' representations as a graph to aggregate the information from similar patients, where the similarity matrix $\Lambda^{\prime}$ is the graph adjacency matrix. We apply graph convolutional layers to enhance the representation learning as:
\begin{equation}
\begin{split}
\label{eq:15}
\hat{e} = [\hat{e}_{1}, \hat{e}_{2}, \cdots, \hat{e}_{B}]^{\top} = GCN(e^{\prime}, \Lambda^{\prime}) \\ = ReLU(\Lambda^{\prime} ReLU(\Lambda^{\prime} \cdot e^{\prime} W^{e}_{1}) \cdot W^{e}_{2}),
\end{split}
\end{equation}
where $\hat{e}$ is the aggregated auxiliary information from similar patients. A note of caution is due here since we ignore the bias term. We replace $e^{\prime}$ in Eq. (15) with $e^{\prime\prime}$ for the imputation task. By doing so, the output of graph convolutional layers can take the form of sequence data. Particularly, $e^{\prime\prime}$ is obtained by concatenating $e$ and $e_{base}$, where $e_{base} \in \mathbb{R}^{d_{g}}$ is expanded into $e_{base} \in \mathbb{R}^{d_{g} \times T}$.

Through the processes above, we are able to generate $e^{\prime}$/$e^{\prime\prime}$ and $\hat{e}$ representations for the batch of patients. The $e^{\prime}$/$e^{\prime\prime}$ refers to the patient themselves. For an incomplete patient $p$ (i.e., the patient data has many missing values), we generate the missing value representations with $\hat{e}$. For a complete patient, we augment $e^{\prime}$/$e^{\prime\prime}$ with $\hat{e}$ to enhance the representation learning.

We design an attention-based fusion module to refine both $e^{\prime}$/$e^{\prime\prime}$ (the two representations used in prediction and imputation tasks) and $\hat{e}$. Since imputation and prediction tasks involve the same process of modeling, we take the prediction task as an example. The two weights $\gamma \in \mathbb{R}^{B}$ and $\eta \in \mathbb{R}^{B}$ are incorporated to determine the importance of $e^{\prime}$ and $\hat{e}$, obtained by implementing fully connected layers as:
\begin{equation}
\begin{split}
\label{eq:17}
\gamma = Sigmoid(e^{\prime} \cdot W_{\gamma} + b_{\gamma}), \\
\eta = Sigmoid(\hat{e} \cdot W_{\eta} + b_{\eta}).
\end{split}
\end{equation}
A note of caution is due here since we keep the sum of $\gamma$ and $\eta$ must be 1, i.e., $\gamma + \eta = 1$. We achieve this constraint by combining $\gamma = \frac{\gamma}{\gamma + \eta}$ and $\eta = 1- \gamma$. The final representation $e^{*}$ is obtained by calculating $\gamma \cdot e^{\prime} + \eta \cdot \hat{e}$.

\subsubsection{Contrastive Learning}
We integrate contrastive learning into the proposed network architecture to perform imputation and prediction tasks. For the prediction task, we augment the standard cross-entropy loss with the supervised contrastive loss \cite{khosla2020supervised}. We treat the patient representations with the same label as the positive pairs and the patient representations with different labels as the negative pairs. For the imputation task,  we augment the standard mean squared error loss with the unsupervised contrastive loss \cite{chen2020simple}. We treat a single patient representation and its augmented representations as positive pairs and the other patient representations within a batch and their augmented representations as negative pairs. The formula can be written as:
\begin{equation}
\begin{split}
\label{eq:16}
\mathcal{L}_{SC} = - \sum_{i = 1}^{B} \frac{1}{B_{y_{i}}}
log \frac{\sum_{j = 1}^{B} \mathbbm{1}_{[y_{i} = y_{j}]} exp(sim(e^{*}_{i}, e^{*}_{j}) / \tau)} {\sum_{k = 1}^{B} \mathbbm{1}_{[k \neq i]} exp(sim(e^{*}_{i}, e^{*}_{k}) / \tau)}, \\
\mathcal{L}_{UC} = - log \frac{exp(sim(e^{*}_{i}, e^{*}_{j}) / \tau)} {\sum^{2B}_{k = 1} \mathbbm{1}_{[k \neq i]} exp(sim(e^{*}_{i}, e^{*}_{k})/\tau)},
\end{split}
\end{equation}
where $B$ represents the batch size; $\mathbbm{1}_{[\cdot]}$ represents an indicator function; $sim(\cdot)$ represents the cosine similarity measure; $\tau$ represents a hyper-parameter that is used to control the strength of penalties on negative pairs; $B_{y_{i}}$ is the number of samples with the same label in each batch.

\subsubsection{Imputation and Prediction Tasks}
For the prediction task, we feed $e^{*}$ into a softmax output layer to obtain the predicted $\hat{y}$ as:
\begin{equation}
\begin{split}
\label{eq:18}
\hat{y} = Softmax(W_{y} \cdot e^{*} + b_{y}).
\end{split}
\end{equation}

The objective loss is the summation of cross-entropy loss and the supervised contrastive loss with a scaling parameter $\lambda$ to control the contribution of each loss as:
\begin{equation}
\begin{split}
\label{eq:19}
\mathcal{L}_{CE} = - \frac{1}{P} \sum^{P}_{p=1} (y_{p}^{\top} \cdot log(\hat{y}_{p}) + (1 - y_{p})^{\top} \cdot log(1 - \hat{y}_{p})), \\
\mathcal{L} = \lambda \cdot \mathcal{L}_{CE} + (1 - \lambda) \cdot \mathcal{L}_{SC}.
\end{split}
\end{equation}

For the imputation task, we take the neighboring observed values (of each patient) as inputs to incorporate patient-specific contextual information. The process of embedding used by $v^{(l)}$ and $v^{(n)}$ can be written as:
\begin{equation}
\begin{split}
\label{eq:20}
e_{i}^{(v), (l)} = FFN_{i}^{(v), (l)} (v_{i}^{(l)}), e_{i}^{(t), (l)} = FFN_{i}^{(t), (l)} (\delta_{i}^{(l)}), \\
e_{i}^{(v), (n)} = FFN_{i}^{(v), (n)} (v_{i}^{(n)}), e_{i}^{(t), (n)} = FFN_{i}^{(t), (n)} (\delta_{i}^{(n)}), \\
\tilde{e}^{(l)} = W_{v}^{(l)} \cdot e^{(v), (l)} + W_{t}^{(l)} \cdot e^{(t), (l)} + b_{e}^{(l)}, \\
\tilde{e}^{(n)} = W_{n}^{(v)} \cdot e^{(v), (n)} + W_{t}^{(n)} \cdot e^{(t), (n)} + b_{e}^{(n)}, \\
e^{c} = Concate(\tilde{e}^{(l)}, \tilde{e}^{(n)}),
\end{split}
\end{equation}
where $\tilde{e}^{(l)}$ and $\tilde{e}^{(n)}$ are the representations of $v^{(l)}$ and $v^{(n)}$ after embedding. The embedding matrix $e^{c}$ is obtained by concatenating $\tilde{e}^{(l)}$ and $\tilde{e}^{(n)}$.

Given the final representation $e^{*}$ and the embedding matrix $e^{c}$, we use a fully connected layer to impute missing values as:
\begin{equation}
\begin{split}
\label{eq:21}
\hat{v} = e^{*} \cdot W^{v}_{1} + e^{c} \cdot W^{v}_{2} + b_{v}.
\end{split}
\end{equation}

The objective loss is the summation of the mean square error and the unsupervised contrastive loss with a scaling parameter $\lambda$ to control the contribution of each loss as:
\begin{equation}
\begin{split}
\label{eq:22}
\mathcal{L}_{MSE} = \frac{1}{P} \sum_{p = 1}^{P} (m_{p} \odot v_{p} - m_{p} \odot \hat{v}_{p})^{2}, \\
\mathcal{L} = \lambda \cdot \mathcal{L}_{MSE} + (1 - \lambda) \cdot \mathcal{L}_{UC}.
\end{split}
\end{equation}

\section{Experiments}
\subsection{Datasets and Tasks}
We validate our approach on the MIMIC-III$\footnote{https://mimic.physionet.org}$ and eICU$\footnote{https://eicu-crd.mit.edu/}$ datasets. We conduct clinical time series imputation and in-hospital mortality experiments based on the data from the first 24/48 hours after admission. Detailed information on both datasets can be found in the literature \cite{johnson2016mimic} and \cite{pollard2018eicu}. The source code of our approach and statistics of features are released at https://github.com/liulab1356/CL-ImpPreNet.
\subsection{Baseline Approaches}
We compare our approach with GRU-D \cite{che2018recurrent}, BRITS \cite{cao2018brits}, GRUI-GAN \cite{luo2018multivariate}, E$^{2}$GAN \cite{luo2019e2gan}, E$^{2}$GAN-RF \cite{zhang2021missing}, STING \cite{oh2021sting}, MTSIT \cite{yildiz2022multivariate}, and MIAM \cite{lee2022multi} (see related work section). We feed the output of GRUI-GAN, E$^{2}$GAN, E$^{2}$GAN-RF, STING, and MTSIT into GRU to estimate in-hospital mortality risk probabilities. Moreover, the regression component used in BRITS is integrated into GRU-D and MIAM to obtain imputation accuracy.

Besides, two variants of our approach are as follows:

Ours$_{\alpha}$: We do not perform graph analysis-based patient stratification modeling.

Ours$_{\beta}$: We omit the contrastive learning component.

All implementations of Ours$_{\alpha}$ and Ours$_{\beta}$ can be found in the aforementioned Github repository.

\subsection{Implementation Details and Evaluation Metrics}
We implement all approaches with PyTorch 1.11.0 and conduct experiments on A40 GPU from NVIDIA with 48GB of memory. We randomly use 70\%, 15\%, and 15\% of the dataset as training, validation, and testing sets. We train the proposed approach using an Adam optimizer \cite{kingma2014adam} with a learning rate of 0.0023 and a mini-batch size of 256. For personalized patient representation learning, the dimension size $d$ is 3. For similar patients discovery and information aggregation, the initial value of $\varphi$ is 0.56, and the dimension size of $W^{e}_{1}$ and $W^{e}_{2}$ are 34 and 55. For contrastive learning, the value of $\tau$ is 0.07. The dropout method is applied to the final Softmax output layer for the prediction task, and the dropout rate is 0.1. For the imputation task, the dimension size of $W^{(l)}_{v}$, $W^{(l)}_{t}$, $W^{(n)}_{v}$, and $W^{(n)}_{t}$ are 28.

The performance of contrastive learning heavily relies on data augmentation. We augment the observed value $v$ with random time shifts and reversion. For example, given the observed value $v = [v_{1}, v_{2}, \cdots, v_{T}]$, we are able to obtain $v_{shift} = [v_{1+n}, v_{2+n}, \cdots, v_{T+n}]$ and $v_{reverse} = [v_{T}, v_{T-1}, \cdots, v_{1}]$ from random time shift and reversion, and $n$ is the number of data points to shift.

We use the MAE and MRE scores between predicted and actual values as the evaluation metrics for imputation performance. We use the AUROC and AUPRC scores as the evaluation metrics for prediction performance. We report the mean and standard deviation of the evaluation metrics after repeating all the approaches ten times.

\section{Experimental Results}
Table 2 presents the experimental results of all approaches on imputation and prediction tasks from MIMIC-III and eICU datasets. Together these results suggest that our approach achieves the best performance in both imputation and prediction tasks. For example, for the clinical time series imputation of MIMIC-III (24 hours after ICU admission), the MAE and MRE of Ours are 0.3563 and 8.16\%, smaller than 0.3988 and 38.44\% achieved by the best baseline (i.e., MTSIT). For the in-hospital mortality prediction of MIMIC-III (24 hours after ICU admission), the AUROC and AUPRC of Ours are 0.8533 and 0.4752, larger than 0.8461 and 0.4513 achieved by the best baseline (i.e., GRU-D).

As Table 2 shows, the RNN-based approach (i.e., GRU-D and BRITS) outperforms the GAN-based approach (i.e., GRUI-GAN, E$^{2}$GAN, E$^{2}$GAN-RF, and STING) in the imputation task. From the prediction results of the MIMIC-III dataset, we can see that the transformer-based approaches (i.e., MTSIT and MIAM) resulted in lower values of AUROC and AUPRC. From the prediction results of the eICU dataset, no significant difference between the transformer-based approach and other approaches was evident.

Ours outperforms its variants Ours$_{\alpha}$ and Ours$_{\beta}$. This result confirms the effectiveness of the network construction with enhanced imputation and prediction performance.

\begin{table*}[htbp] \scriptsize
  \centering
  \caption{Performance of our approaches with other baselines on clinical time series imputation and in-hospital mortality prediction.} 
    \begin{tabular}{crrcc}
    \toprule
    MIMIC-III/24 hours after ICU admission & \multicolumn{2}{c}{Clinical time series imputation} & \multicolumn{2}{c}{In-hospital mortality prediction} \\
    \midrule
    Metrics & \multicolumn{1}{c}{MAE} & \multicolumn{1}{c}{MRE} & AUROC & AUPRC \\
    \midrule
    GRU-D & \multicolumn{1}{c}{1.3134(0.0509)} & \multicolumn{1}{c}{87.33\%(0.0341)} & 0.8461(0.0051) & 0.4513(0.0124) \\
    BRITS & \multicolumn{1}{c}{1.3211(0.0923)} & \multicolumn{1}{c}{87.92\%(0.0611)} & 0.8432(0.0040) & 0.4193(0.0144) \\
    GRUI-GAN & \multicolumn{1}{c}{1.6083(0.0043)} & \multicolumn{1}{c}{107.20\%(0.0029)} & 0.8324(0.0077) & 0.4209(0.0280) \\
    E$^{2}$GAN & \multicolumn{1}{c}{1.5885(0.0045)} & \multicolumn{1}{c}{105.86\%(0.0032)} & 0.8377(0.0083) & 0.4295(0.0137) \\
    E$^{2}$GAN-RF & \multicolumn{1}{c}{1.4362(0.0031)} & \multicolumn{1}{c}{101.09\%(0.0027)} & 0.8430(0.0065) & 0.4328(0.0101) \\
    STING & \multicolumn{1}{c}{1.5018(0.0082)} & \multicolumn{1}{c}{102.53\%(0.0047)} & 0.8344(0.0126) & 0.4431(0.0158) \\
    MTSIT & \multicolumn{1}{c}{0.3988(0.0671)} & \multicolumn{1}{c}{38.44\%(0.0647)} & 0.8029(0.0117) & 0.4150(0.0165) \\
    MIAM  & \multicolumn{1}{c}{1.1391(0.0001)} & \multicolumn{1}{c}{75.65\%(0.0001)} & 0.8140(0.0044) & 0.4162(0.0079) \\
    Ours  &  \multicolumn{1}{c}{\textbf{0.3563(0.0375)}} & \multicolumn{1}{c}{\textbf{8.16\%(0.0086)}} & \textbf{0.8533(0.0119)} & \textbf{0.4752(0.0223)} \\
    Ours$_{\alpha}$ & \multicolumn{1}{c}{0.3833(0.0389)} & \multicolumn{1}{c}{8.78\%(0.0089)} & 0.8398(0.0064) & 0.4555(0.0139) \\
    Ours$_{\beta}$ & \multicolumn{1}{c}{0.4125(0.0319)} & \multicolumn{1}{c}{8.95\%(0.0077)} & 0.8417(0.0059) & 0.4489(0.0182) \\
    \midrule
    eICU/24 hours after eICU admission  & \multicolumn{2}{c}{Clinical time series imputation} & \multicolumn{2}{c}{In-hospital mortality prediction} \\
    \midrule
    Metrics & \multicolumn{1}{c}{MAE} & \multicolumn{1}{c}{MRE} & AUROC & AUPRC \\
    \midrule
    GRU-D & \multicolumn{1}{c}{3.9791(0.2008)} & \multicolumn{1}{c}{52.11\%(0.0262)} & 0.7455(0.0107) & 0.3178(0.0190) \\
    BRITS & \multicolumn{1}{c}{3.6879(0.3782)} & \multicolumn{1}{c}{48.30\%(0.0726)} & 0.7139(0.0101) & 0.2511(0.0111) \\
    GRUI-GAN & \multicolumn{1}{c}{9.1031(0.0130)} & \multicolumn{1}{c}{119.29\%(0.0016)} & 0.7298(0.0094) & 0.3013(0.0141) \\
    E$^{2}$GAN & \multicolumn{1}{c}{7.5746(0.0141)} & \multicolumn{1}{c}{99.20\%(0.0018)} & 0.7317(0.0155) & 0.2973(0.0253) \\
    E$^{2}$GAN-RF & \multicolumn{1}{c}{6.7108(0.0127)} & \multicolumn{1}{c}{90.38\%(0.0015)} & 0.7402(0.0131) & 0.3045(0.0227) \\
    STING & \multicolumn{1}{c}{7.1447(0.0651)} & \multicolumn{1}{c}{93.56\%(0.0083)} & 0.7197(0.0154) & 0.2873(0.0182) \\
    MTSIT & \multicolumn{1}{c}{1.6192(0.1064)} & \multicolumn{1}{c}{21.20\%(0.0139)} & 0.7215(0.0071) & 0.2992(0.0115) \\
    MIAM  & \multicolumn{1}{c}{1.1726(0.3103)} & \multicolumn{1}{c}{15.35\%(0.0406)} & 0.7262(0.0179) & 0.2659(0.0148) \\
    Ours  & \multicolumn{1}{c}{\textbf{0.5365(0.0612)}} & \multicolumn{1}{c}{\textbf{7.02\%(0.0079)}} & \textbf{0.7626(0.0117)} & \textbf{0.3388(0.0211)} \\
    Ours$_{\alpha}$ & \multicolumn{1}{c}{0.6792(0.0716)} & \multicolumn{1}{c}{8.89\%(0.0093)} & 0.7501(0.0143) & 0.3325(0.0151) \\
    Ours$_{\beta}$ & \multicolumn{1}{c}{0.5923(0.0514)} & \multicolumn{1}{c}{7.75\%(0.0067)} & 0.7533(0.0104) & 0.3303(0.0175) \\
    \midrule
    MIMIC-III/48 hours after ICU admission & \multicolumn{2}{c}{Clinical time series imputation} & \multicolumn{2}{c}{In-hospital mortality prediction} \\
    \midrule
    Metrics & \multicolumn{1}{c}{MAE} & \multicolumn{1}{c}{MRE} & AUROC & AUPRC \\
    \midrule
    GRU-D & \multicolumn{1}{c}{1.4535(0.0806)} & \multicolumn{1}{c}{86.47\%(0.0482)} & 0.8746(0.0026) & 0.5143(0.0077) \\
    BRITS & \multicolumn{1}{c}{1.3802(0.1295)} & \multicolumn{1}{c}{82.21\%(0.0768)} & 0.8564(0.0040) & 0.4445(0.0189) \\
    GRUI-GAN & \multicolumn{1}{c}{1.7523(0.0030)} & \multicolumn{1}{c}{104.50\%(0.0018) } & 0.8681(0.0077) & 0.5123(0.0166) \\
    E$^{2}$GAN & \multicolumn{1}{c}{1.7436(0.0036)} & \multicolumn{1}{c}{103.98\%(0.0022)} & 0.8705(0.0043) & 0.5091(0.0120) \\
    E$^{2}$GAN-RF & \multicolumn{1}{c}{1.6122(0.0027)} & \multicolumn{1}{c}{102.34\%(0.0017)} & 0.8736(0.0031) & 0.5186(0.0095) \\
    STING & \multicolumn{1}{c}{1.6831(0.0068)} & \multicolumn{1}{c}{100.46\%(0.0035)} & 0.8668(0.0123) & 0.5232(0.0236) \\
    MTSIT & \multicolumn{1}{c}{0.4503(0.0465)} & \multicolumn{1}{c}{30.42\%(0.0314)} & 0.8171(0.0114) & 0.4308(0.0189) \\
    MIAM  & \multicolumn{1}{c}{1.3158(0.0003)} & \multicolumn{1}{c}{78.20\%(0.0002)} & 0.8327(0.0024) & 0.4460(0.0061) \\
    Ours  &  \multicolumn{1}{c}{\textbf{0.4396(0.0588)}} & \multicolumn{1}{c}{\textbf{6.23\%(0.0073)}} & \textbf{0.8831(0.0149)} & \textbf{0.5328(0.0347)} \\
    Ours$_{\alpha}$ & \multicolumn{1}{c}{0.7096(0.0532)} & \multicolumn{1}{c}{8.85\%(0.0066)} & 0.8671(0.0093) & 0.5161(0.0151) \\
    Ours$_{\beta}$ & \multicolumn{1}{c}{0.5786(0.0429)} & \multicolumn{1}{c}{7.47\%(0.0056)} & 0.8709(0.0073) & 0.5114(0.0176) \\
    \midrule
    eICU/48 hours after eICU admission  & \multicolumn{2}{c}{Clinical time series imputation} & \multicolumn{2}{c}{In-hospital mortality prediction} \\
    \midrule
    Metrics & \multicolumn{1}{c}{MAE} & \multicolumn{1}{c}{MRE} & AUROC & AUPRC \\
    \midrule
    GRU-D & \multicolumn{1}{c}{5.8071(0.2132)} & \multicolumn{1}{c}{44.53\%(0.0164)} & 0.7767(0.0141) & 0.3210(0.0182) \\
    BRITS & \multicolumn{1}{c}{5.5546(0.5497)} & \multicolumn{1}{c}{42.59\%(0.0421)} & 0.7285(0.0114) & 0.2510(0.0097) \\
    GRUI-GAN & \multicolumn{1}{c}{14.0750(0.0301)} & \multicolumn{1}{c}{107.96\%(0.0021)} & 0.7531(0.0167) & 0.2897(0.0201) \\
    E$^{2}$GAN & \multicolumn{1}{c}{12.9694(0.0195)} & \multicolumn{1}{c}{99.47\%(0.0015)} & 0.7605(0.0063) & 0.3014(0.0137) \\
    E$^{2}$GAN-RF & \multicolumn{1}{c}{11.8138(0.0161)} & \multicolumn{1}{c}{91.52\%(0.0011)} & 0.7763(0.0057) & 0.3101(0.0125) \\
    STING & \multicolumn{1}{c}{12.0962(0.0806)} & \multicolumn{1}{c}{92.79\%(0.0062)} & 0.7453(0.0182) & 0.2805(0.0190) \\
    MTSIT & \multicolumn{1}{c}{2.8150(0.2105)} & \multicolumn{1}{c}{21.58\%(0.0161)} & 0.7418(0.0091) & 0.3078(0.0120) \\
    MIAM  & \multicolumn{1}{c}{2.1146(0.4012)} & \multicolumn{1}{c}{16.23\%(0.0414)} & 0.7574(0.0127) & 0.2776(0.0105) \\
    Ours  & \multicolumn{1}{c}{\textbf{0.9412(0.0930)}} & \multicolumn{1}{c}{\textbf{7.21\%(0.0071)}} & \textbf{0.7907(0.0123)} & \textbf{0.3417(0.0217)} \\
    Ours$_{\alpha}$ & \multicolumn{1}{c}{1.1099(0.1064)} & \multicolumn{1}{c}{8.51\%(0.0081)} & 0.7732(0.0100) & 0.3311(0.0265) \\
    Ours$_{\beta}$ & \multicolumn{1}{c}{0.9930(0.0817)} & \multicolumn{1}{c}{7.61\%(0.0062)} & 0.7790(0.0117) & 0.3335(0.0178) \\
    \bottomrule
    \end{tabular}%
  \label{tab:addlabel}%
\end{table*}%

\section{Conclusion}
This paper presents a novel contrastive learning-based imputation-prediction network to carry out in-hospital mortality prediction tasks using EHR data. This prediction makes timely warnings available to ICU health professionals so that early interventions for patients at risk could take place.
The proposed approach explicitly considers patient similarity by stratification of EHR data and successfully integrates contrastive learning into the network architecture. We empirically show that the proposed approach outperforms all the baselines by conducting clinical time series imputation and in-hospital mortality prediction on the MIMIC-III and eICU datasets.

\section{Acknowledgement}
This research is partially funded by the ARC Centre of Excellence for Automated Decision-Making and Society (CE200100005) by the Australian Government through the Australian Research Council.

\bibliographystyle{splncs04}
\bibliography{BIB}
\end{document}